\documentclass[letterpaper, 10 pt, conference]{ieeeconf}  

\IEEEoverridecommandlockouts                              

\usepackage{amsmath,amsfonts}
\usepackage[linesnumbered,ruled,vlined]{algorithm2e}
\usepackage{algorithmic}
\usepackage{graphicx}
\usepackage{textcomp}
\usepackage{xcolor}
\usepackage{caption} 
\usepackage{subfigure}
\usepackage{amsmath}
\usepackage[normalem]{ulem}
\usepackage{blindtext}
\usepackage{grffile}

\title{\LARGE \bf

Improving Responsiveness to Robots for Tacit Human-Robot Interaction \\ via Implicit and Naturalistic Team Status Projection 
}

\author{Andrew Boateng, Wenlong Zhang and Yu Zhang
\thanks{A. Boateng and Y. Zhang are with the 
School of Computing and Augmented Intelligence, Ira A. Fulton Schools of Engineering, Arizona State University, Tempe, AZ 85281, USA. Email: {\tt\small $\{$aoboaten, yu.zhang.442$\}$@asu.edu}.}
\thanks{W. Zhang is with the School of Manufacturing Systems and Networks, Ira A. Fulton Schools of Engineering, Arizona State University, Mesa, AZ 85212, USA. Email: {\tt\small wenlong.zhang@asu.edu}.}}

\begin{document}

\maketitle
\thispagestyle{empty}
\pagestyle{empty}

\begin{abstract}
Fluent human-human teaming is often characterized by tacit interaction without explicit communication. 
This is because explicit communication, such as language utterances and gestures, are inherently interruptive. 
On the other hand, tacit interaction requires team situation awareness (TSA) to facilitate,
which often relies on explicit communication to maintain, creating a paradox. 
In this paper, we consider implicit and naturalistic team status projection for tacit human-robot interaction. 
Implicitness minimizes interruption while naturalness reduces cognitive demand, and they together improve responsiveness to robots. 
We introduce a novel process for such {\textit{T}}eam status {\textit{P}}rojection via virtual {\textit{S}}hadows, or {\it TPS}. 
We compare our method with two baselines that use explicit projection for maintaining TSA.
Results via human factors studies demonstrate that TPS provides a more fluent human-robot interaction experience by significantly improving human responsiveness to robots in tacit teaming scenarios, which suggests better TSA. 
Participants acknowledged robots implementing TPS as more acceptable as a teammate and favorable.
Simultaneously, we demonstrate that TPS is comparable to, and sometimes better than, the best-performing baseline in maintaining accurate TSA.  
\end{abstract}

\section{INTRODUCTION}
Over the past decade, there has been accelerated growth and advancement in robotic search, making it no longer far-fetched to envision robots as part of our lives. One of the most appealing applications are teaming domains where humans and robots complement each other to achieve complex tasks ~\cite{OLIFF}.
Effective teaming is often characterized by tacit interaction without explicit communication\footnote{In our work, 
we take a more generic stance and refer to explicit communication as communication, regardless of its modality, with the intention to convey information {\it and} be perceived as conveying information.} to minimize interruption. 
To facilitate such interaction, 
it is critical for the team members to maintain team situation awareness (TSA)   where each member separately maintains and predicts the team status for non-interruptive and fluent teaming  ~\cite{nancy}. 
However, existing methods for human-robot teaming (HRT) often rely on explicit communication for maintaining TSA~\cite{mavridis2015review, gleeson2013gestures, fong2003collaboration},
creating a paradox for achieving tacit human-robot interaction (HRI).
While implicit communication methods have been investigated mainly on communicative robot motion and actions~\cite{cohen1995communicative, dragan2013legibility, zhang2017plan}, 
these methods would only work under the assumption of continuous observation of the robot and its actions.
Hence, the challenge of maintaining TSA {\it without explicit communication} is still left unattended.

Consider a scenario in a semi-automated car assembly shop where a human worker, Mark, works along with a partner robot. Each agent has its own tasks in hand but must also collaborate occasionally to make progress. In one scenario, Mark sends the robot to fetch a hot soldering rod. 
However, Mark may not know exactly when the robot would return. To improve productivity, he would context switch to others tasks before the robot returns instead of idly waiting for the rod. 
However, focusing on the other tasks would cause him to lose track of the moving robot, resulting in the loss of TSA (e.g., whether the robot is approaching), 
degraded team performance,  and even safety risks. 
Similarly, having the robot announcing its arrival can be interruptive and does not help much with continuous TSA maintenance other than providing discretized status-updates. 
In such situations, it would be desirable for the robot to {\it implicitly project} its status to Mark in a way that draws little attention to allow him to continuously track the robot while focusing on the other tasks. 
Simultaneously, we would like to encourage Mark to be responsive to the robot to reduce its waiting time as if with a human teammate, for improved team efficiency.

\begin{figure}[t!]
    \centering
  \includegraphics[scale=0.14]{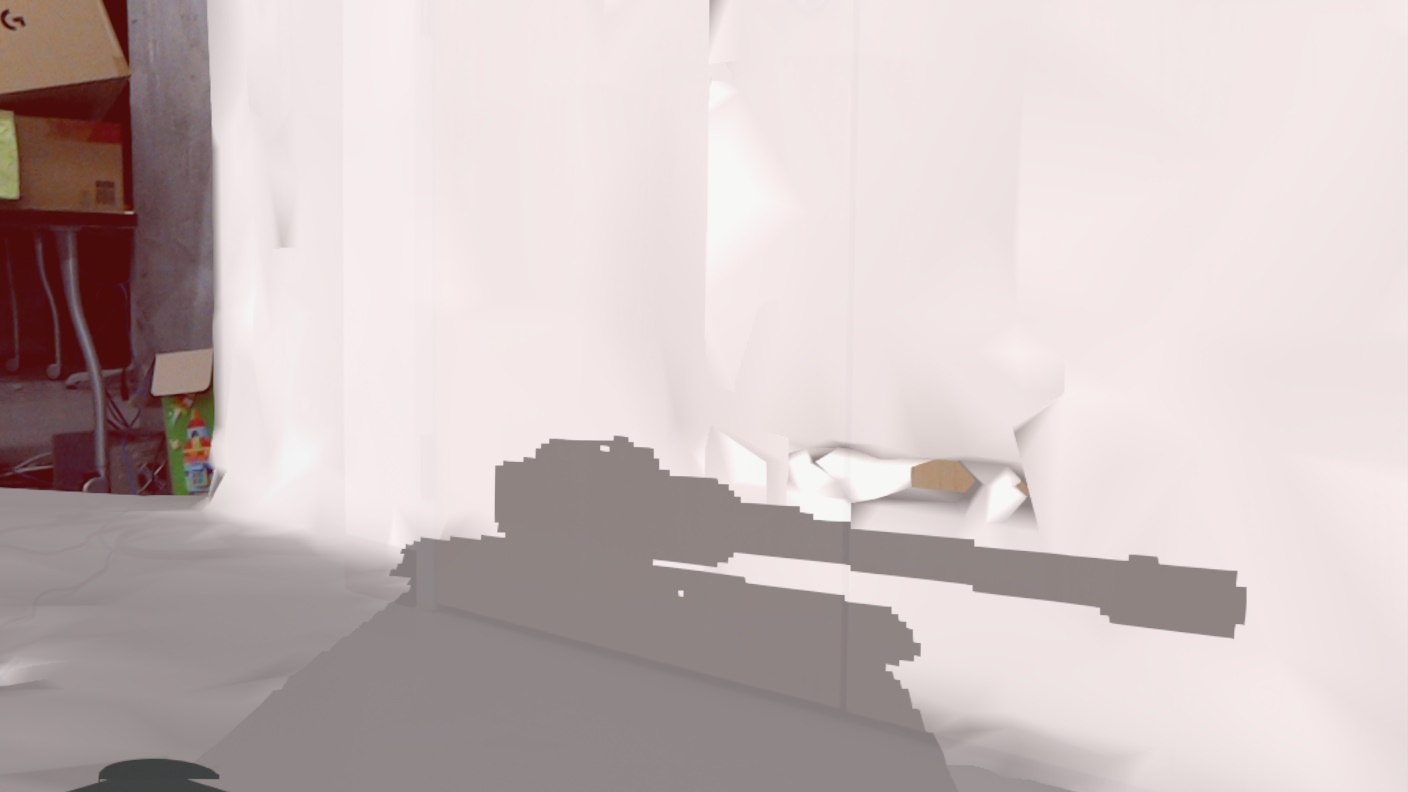}
  \caption{A virtual shadow of a 3D tank model generated by TPS on a flat surface connected against a ``wall''}
  \label{fig:motivation}
 \vskip-10pt
\end{figure}

In this paper, we consider implicit and naturalistic \textbf{T}eam status \textbf{P}rojection via virtual \textbf{S}hadows (TPS). 
TPS is considered implicit communication in our work since shadows are not normally {\it perceived} as a way for conveying information. 
Implicit projection minimizes interruption while naturalness reduces cognitive demand, and they together improve human responsiveness to robots. 
In particular, the virtual shadows are created in a way to simulate the  experience in proximal human-human teaming where observing the shadows of others would help us infer about their status. It has two intuitive appeals given our familiarity with shadows: 1) we can easily and implicitly interpret the shadows of others; 2) the shadows of agents around us would naturally encourage our interaction with them (and hence our responsiveness to them as well)~\cite{reis2011familiarity}. 
To realize TPS, we propose to use Augmented Reality (AR) to generate virtual shadows.
However, there are significant scientific and engineering challenges to address: 
\begin{itemize}

    \item {\it Realistic Shadow Projection}: Shadow projection is often achieved with approximate methods such as negative shadow. However, for proximal human-robot interaction, unrealistic shadow projection can lead to misinterpretation of the shadow and hence TSA. 
    
    \item {\it Naturalistic Shadow Projection}: To display shadows when needed, the virtual light source must be dynamically updated to project the shadow into the human's field of view. This can have a negative effect on the perception of the naturalness of virtual shadows, leading to misinterpretation and unwarranted distraction. 
    
    
\end{itemize}

Our contributions in this work is three-fold. T First, we developed a novel engineering process for realistic shadow projection with Microsoft Hololens, which includes  environment modeling and virtual shadow rendering\footnote{This part was presented as a Late-Breaking Report at HRI 2021~\cite{boateng2021virtual}.}. 
It generates shadows by superimposing a cutout model from 3D scan of the environment for generating the robot's shadow onto the real-world. 
The result is high fidelity shadows generated according to the environment layout as with real shadows. 
See Fig. \ref{fig:motivation} for an illustration of a virtual shadow generated by TPS. 
Second, we proposed a method for naturalistic shadow projection to ensure that the shadows are  {\it informative} and {\it smooth}, by integrating a virtual shadow mapping mechanism with a control method. 
The shadow mapping mechanism ensures that the robot status (such as its moving and rotation speeds) is ``effectively'' represented and the control method makes sure that  shadow generation is smooth to minimize discomfort during interaction. 
Third, we evaluated TPS to validate our hypotheses and compared with baselines (that use explicit projection) to demonstrate its effectiveness in facilitating tacit teaming for proximal HRI. 

\section{RELATED WORK}
AR empowers us to visually perceive and interact with objects that are not present in the physical world ~\cite{aph_144559232}. Due to the intuitive appeal of such visual augmentation,
it has been used in various domains ~\cite{Mekni2014AugmentedR} that include military ~\cite{liv}, marketing ~\cite{marketing}, education ~\cite{MustafaS}, manufacturing~\cite{engineering}, medical~\cite{medical}, entertainment~\cite{entertainment}, and robotics~\cite{robot}. 
Thus far, research has been focused on making VR objects more realistic and interactable. 
For example, Wang et. al ~\cite{WANG2003185} make use of the lighting and shading of real scenes to modify AR objects to make them more lifelike.
In robotics, AR objects have been used as part of the interface to facilitate human-robot interaction ~\cite{control}. 
Anderson et al. ~\cite{heni} project virtual parts onto a physical object to highlight the right places to insert the parts for assembly tasks.
In general, AR has been mostly used as an explicit way of visual communication,
which often requires substantial human attention. 
Our focus in this work is to consider {\it implicit communication} with AR. 


In linguistics, traditionally, explicit communication refers to information conveyed via spoken or written words. 
When multiple modalities are available, explicit communication may be expanded to refer to information conveyed directly through an established channel while implicit communication to information that must be inferred. 
Such a distinction, however, is subject to a level of ambiguity.  
For example, spoken words may also have implied meanings that are not directly available and must be inferred ~\cite{carston2008thoughts, carston2009explicit}.   
In our work, 
we take a more generic stance and refer to explicit communication, regardless of its modality, as communication with the intention to convey information {\it and} be perceived as conveying information. 
Note that our definition of explicit communication is similar to that commonly adopted in HRI research ~\cite{breazeal2005effects, che2020efficient}, 
except that we also consider it {\it from the receiver's perspective}.
While implicit communication methods have been investigated in HRI, mainly on communicative robot motion and actions~\cite{cohen1995communicative, dragan2013legibility, zhang2017plan},
they would only work under the assumption of continuous human observation. 
TPS specifically addresses the challenge of maintaining TSA without explicit communication for proximal HRI using AR. 



\section{APPROACH}

\begin{figure}[t!]
    \centering
  \includegraphics[width=0.8\linewidth]{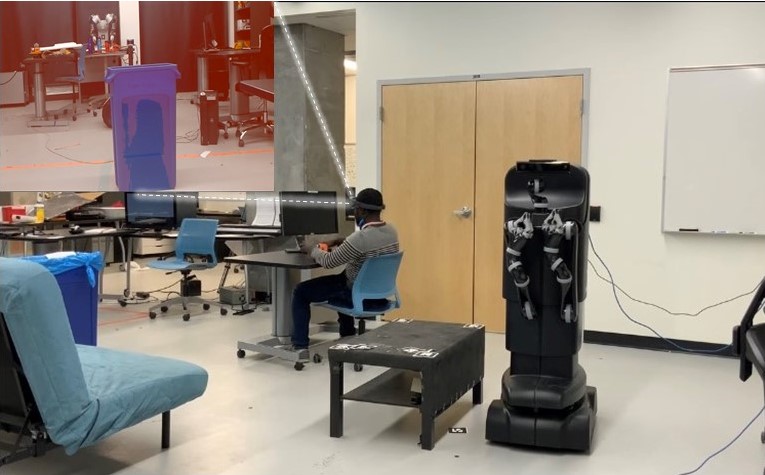}
  \caption{An example of task settings for TPS in this paper where the top left shows the user view through HoloLens}
  \label{task}
 \vskip-10pt
\end{figure}

To maintain TSA, 
TPS requires the virtual shadow to be always observable to the human. 
To simplify the technical development, 
we assume that the robot would always operate behind the human (i.e., outside of the human's field of view) and the human would not change his viewing directions.
For example, this situation may occur when the human is reading off a computer. 
See Fig. \ref{task} for an example of the task settings used in this paper. 
The relaxation of these assumptions will be discussed in future work.


\subsection{Shadow Mapping}
\label{sec:mapping}

\begin{figure}[h!]
\centerline{\includegraphics[width=0.8\linewidth]{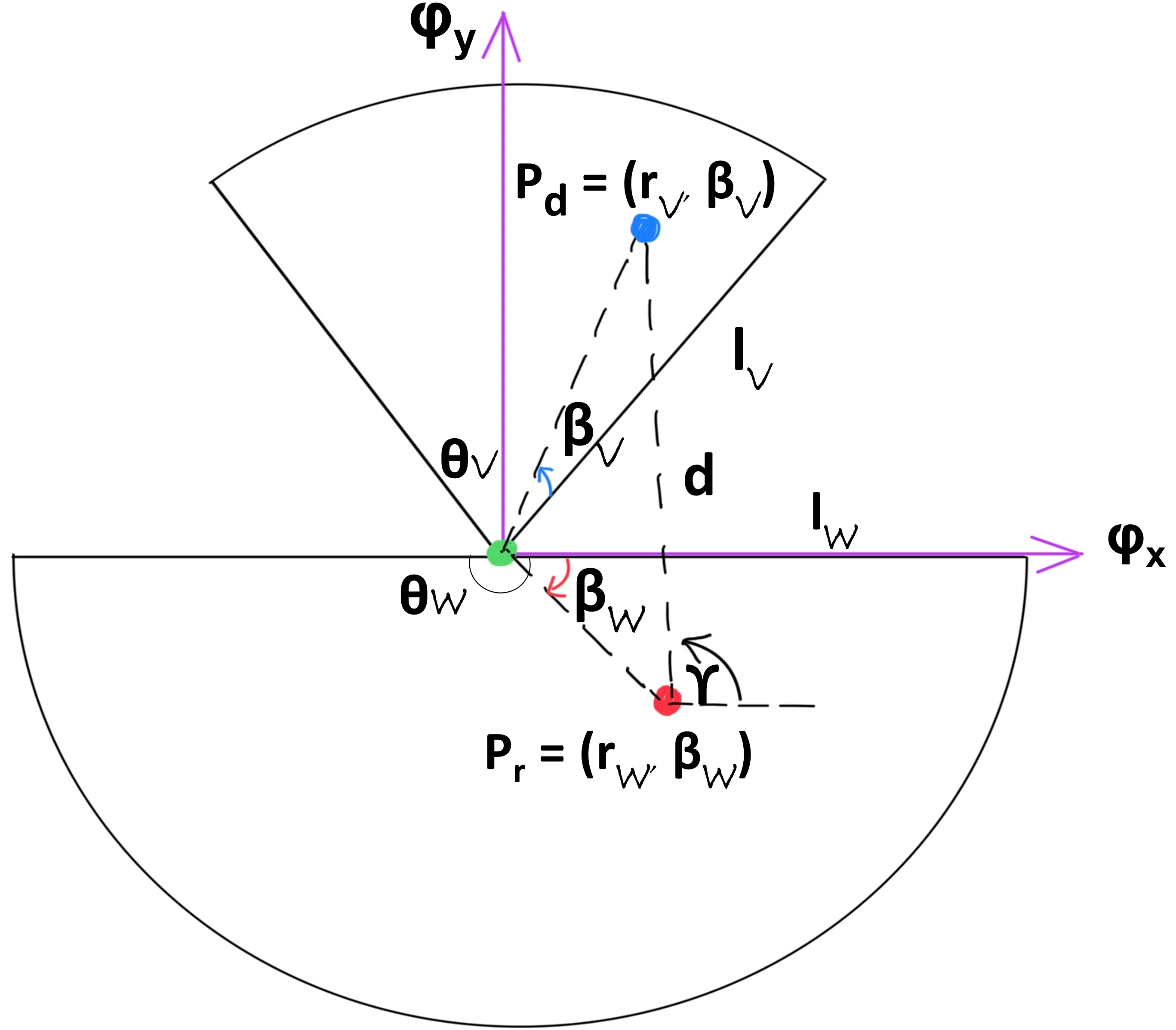}}
 \caption{Projection from the robot's position ${\color{purple}\left(\color{purple}r_w, \beta_w\right)}$ in the real-world to its shadow position (i.e., the top of the robot's shadow) in the virtual world ${\color{blue}\left(r_v, \beta_v\right)}$}
 \label{mapping}
 \vskip-10pt
\end{figure}

One of the challenges to render the virtual shadow always observable is that the Hololens has a small field of view (FOV) of $34^{\circ}$ with a maximum distance of $5m$ from user to the holograms (i.e., AR objects). 
To achieve this, 
we use {\it Shadow Mapping} to project the robot's position from outside the human's FOV in the real-world  to its desired shadow position in the virtual world within the FOV of Hololens.
Furthermore, to ensure that the shadow is {\it informative} about the robot's status, 
we would like the shadow movements to effectively capture the movements of the robot, given that the exactly correct mapping (as with real shadows) may not always result in a visible shadow. 
Intuitively, to ensure its effectiveness, when the robot moves faster (slower), the shadow should also move faster (slower); for sufficiently small enough position updates, when the robot moves left, right, up, or down, the shadow should also move likewise.
To satisfy these requirements, we choose to implement a linear mapping between the robot's position and the shadow's position in their respective polar coordinate systems. 
First, we consider the real-world outside the human's FOV  to be a semi-circular area (i.e., $\theta_w = 180^{\circ}$) with a user declared radius $l_w$ (i.e., the maximum distance from the human to the robot where the robot's status is critical to the human), and the virtual world as a sector with apex angle $\theta_v = 34^{\circ}$  and $l_v = 5m$. 
The mapping is as follows:

\vskip-10pt
\begin{eqnarray}
    {\color{blue}r_v} = {\color{blue}l_v} - {\color{purple}r_w}\frac{\color{blue}{l_v}}{\color{purple}l_w}, \,\,\,\,\beta_v = {\color{purple}\beta_w}\frac{{\color{blue}\theta_v}}{\color{purple}{\theta_w}},
\label{eq:mapping}
\end{eqnarray}
%
where 
$r_w$ and $\beta_w$ above refer to the polar coordinates of a point in the real-world, and $r_v$ and $\beta_v$ refer to the polar coordinates of the corresponding point in the virtual world. 
Note that $r_v$ is a decreasing function of $r_w$ since more shadow should be seen when the robot moves closer to the human. 
Such a mapping is illustrated in Fig. \ref{mapping} where the human is at the intersection of the two worlds shown as a green dot.
We also introduce the global coordinate system as a Cartesian system (i.e., $\varphi_{x}$ and $\varphi_{y}$). 

 \subsection{Shadow Projection}
 \label{sec:projection}
\begin{figure}[htbp]
\centerline{\includegraphics[width=0.7\linewidth]{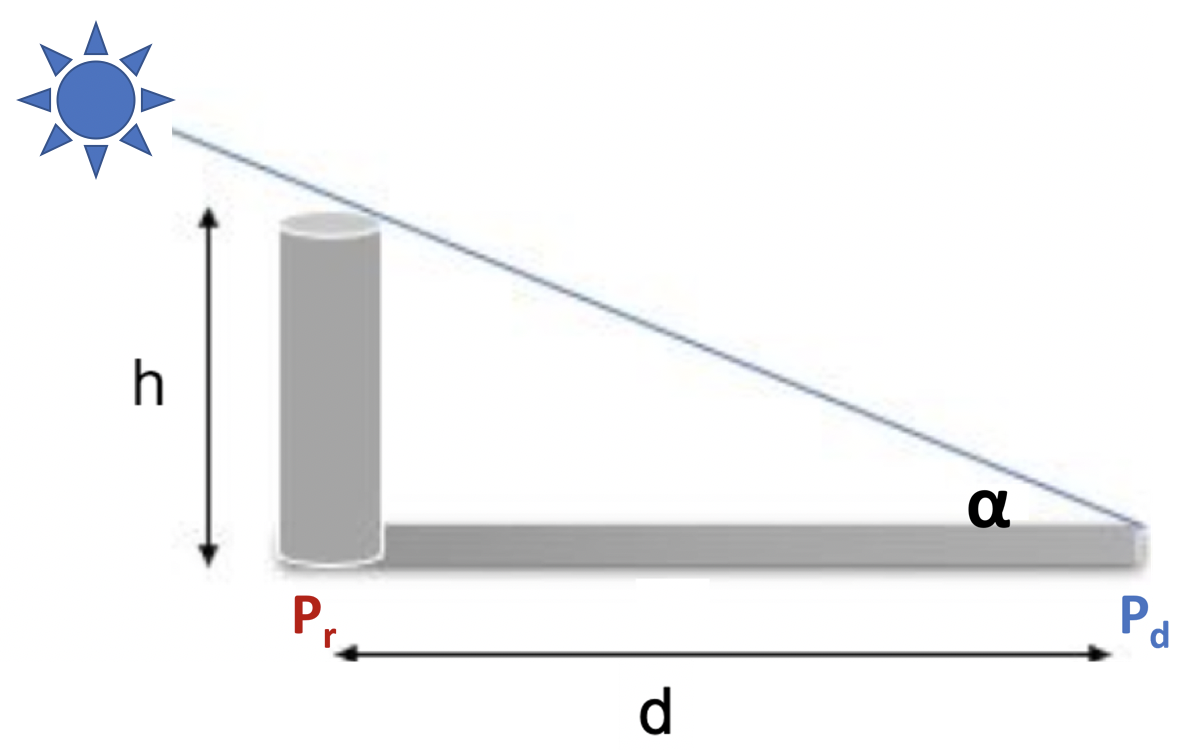}}
\vskip-10pt
 \caption{Relationship between the tilt $\alpha$ of the  light source, robot height, $h$, and shadow-robot distance ($d$)}
 \label{sun}
\end{figure}

 Next, we discuss how to project the robot's position, denoted as $P_r$, to its desired shadow position as expressed in Eq. \eqref{eq:mapping}, denoted as  $P_d$, by setting the tilt and pan of a directional light source. 
The 3D development platform (Unity) for Hololens uses a depth buffer system to keep track of all surfaces close to the light source. If any surface comes in direct line with the light source, the surface will be illuminated. The unilluminated surface therefore creates the shadow effect~\cite{unity}. 
The benefit of using such a process is so that the shadow generated will be realistic as it naturally caters to the surface onto which the shadow is projected (see Fig. \ref{Transp}), assuming that the environment model is accurate.  
Hence, we can safely ignore the geometry of the virtual world in shadow projection by assuming that it is a flat surface.
For a given $P_d$ and height (\emph{h}) of the robot, the tilting, $\alpha$, of the light source  to generate a shadow long enough to reach $P_d$ is given below and illustrated in Fig. \ref{sun}: 
\begin{equation}
\label{ele}
    \alpha = \tan^{-1}\left(\frac{h}d\right)
\end{equation} 

\vskip-20pt
Adjusting the tilt of the light source would increase or decrease the shadow length as needed.
$d$ is 
the Euclidean distance between  $P_r = \color{purple}\left(r_w, \beta_w\right)$ and $P_d = {\color{blue}(r_v, \beta_v)}$:
\begin{equation}
\label{euclidean}
    d = euclidean({\color{purple}\left(r_w, \beta_w\right)}, {\color{blue}(r_v, \beta_v)})
\end{equation}
We compute the pan ($\gamma$) of the light source based on Fig. \ref{mapping}. 

\subsection{Shadow Smoothing}
Even though we can derive the exact tilt and pan of the light source to project the top of the robot to the desired shadow position as discussed above, 
the magnitudes of the updates to these angles for when the robot moves in different parts of the real world can differ substantially.
This means significant variation in the angle updates.
For example, for the same amount of shadow movement, the smaller the shadow-robot distance (i.e., $d$) is, the more the light source must update its tilt (see Eq. \eqref{ele}). 
Users are often not accustomed to significant directional changes of the light source, 
which could mislead the perception of the shadow and the maintenance of TSA. 
We must reduce such effects by smoothing the shadow generation process. 

We choose to apply a PID control method that is often used in robotics to generate smoother state transition processes~\cite{PIDforRos}. 
It is a combination of Proportional ($P$), Integral ($I$), and Derivative ($D$) controls. \emph{P} is proportional to the error between a set point and the observed process variable.
The term \emph{I} considers the past errors and integrates them over time to help correct the accumulated error. The \emph{D} term estimates the future error. The control function of PID is given by:
\begin{equation}
    u(k) = K_{p}e(k) + K_{i}\sum_{\tau = 0}^{k} e(\tau) + K_{d}(e(k) - e(k-1))
\end{equation}
where $K_{p}, K_{i}$ and $K_{d}$ are the coefficient $2 \times 2$ matrices of $P$, $I$ and $D$, respectfully.
For shadow smoothing, 
with changes to the light source angles (i.e., $u$) as our control inputs and given the robot's position in the real world (i.e., $P_r$), we must drive the shadow towards the desired output $P_d$. 
Such a model can be modeled with the plant as follow:
\begin{equation} 
    x(k+1) = f(x(k), u(k), \Delta P_r(k)),
    \label{p}
\end{equation} 
where $x$ is the virtual shadow position, and $u = [\Delta \alpha,\Delta \gamma]^T$ encodes changes to the tilt and pan angles of the light source that we are actively controlling.  {\color{black} $\Delta P_r$ is the change in the robot's position in the real world, 
which is treated as an {\color{black}exogenous} input. 
} 
In this paper, we consider $f$ as a {\color{black}first-order discrete-time dynamic model}:
\begin{equation} 
    x(k+1) = x(k) + 
    \begin{bmatrix} 
	-a & 0 \\
	0 & b \\
    \end{bmatrix}u(k) + \begin{bmatrix} 
	-f & 0 \\
	0 & g \\
    \end{bmatrix}\Delta P_r(k)
\end{equation} 
where $a$, $b$, $f$, and $g$ are positive constants.
These values are chosen to capture $\Delta P_r(k)$ and $u(k)$'s expected relationship with the change of the shadow position from step $k$ to $k + 1$. 
{\color{black}We assume that $\Delta P_r(k)$ is expressed in the polar coordinate system of the real world.}



Now, we can derive a simple PID controller using Eq. \eqref{p} with the setpoint at step $k$ being $P_d(k)$. This is the position we would like the shadow to be rendered. {\color{black} ${x(k)}$ represents the shadow position actually rendered at step $k$. We assume that both ${x(k)}$ and $P_d(k)$ are expressed in the polar coordinate system of the virtual world. The difference between $P_d(k)$ and ${x(k)}$ then leads to the error $e(k) = P_d(k) - {x(k)}$.}


\subsection{Shadow Rendering}
To generate realistic shadows in TPS, 
shadow rendering is composed of environment modeling,  shadow generation, and shadow superimposition. 
Our environment modeling technique uses the semi-autonomous nature of SLAM-like modeling (provided by HoloLens). 
To be able to find anchoring surfaces in the real world to place virtual objects (Holograms), the HoloLens constantly maps its environment. This also ensures that when there is a change in the environment (e.g., when an object is moved in the environment), it will be updated to the new arrangement. 
\begin{figure}[htbp]
\centerline{\includegraphics[width=0.8\linewidth]{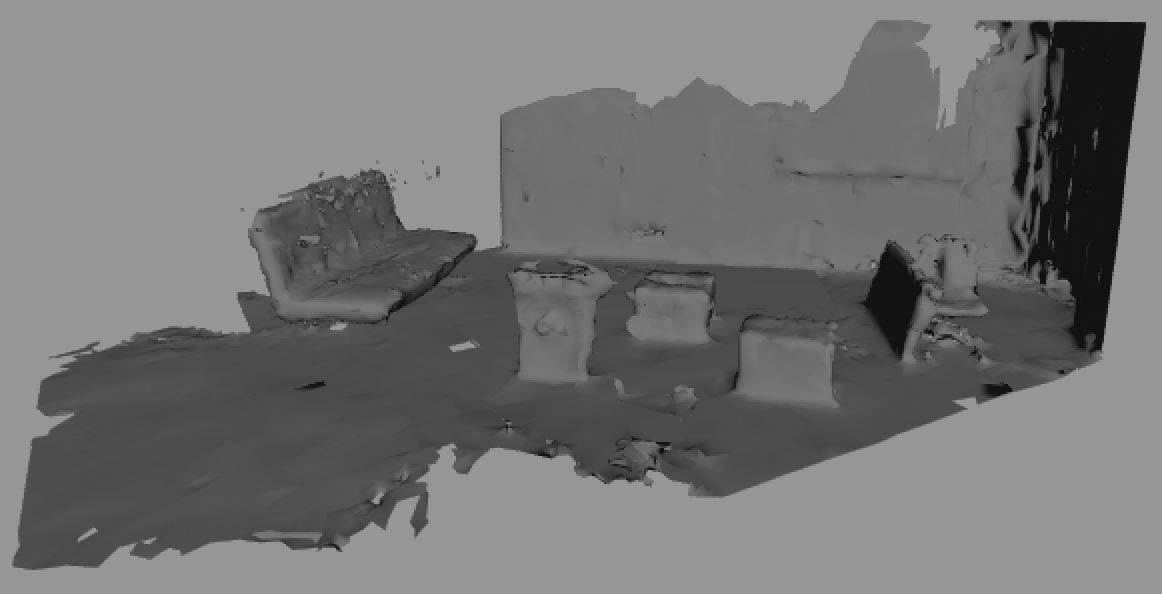}}
 \caption{3D scan by Hololens for environment modeling}
 \label{2ndFloor}
\vskip-10pt
\end{figure}

\begin{figure}[htbp]
\centerline{\includegraphics[width=0.8\linewidth]{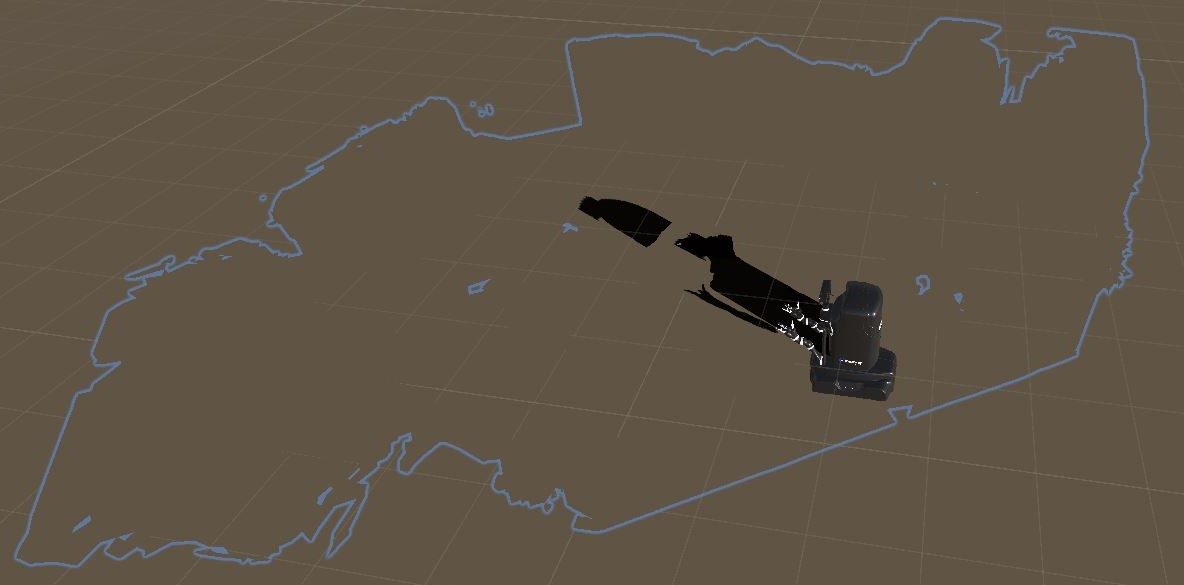}}
 \caption{A transparent shadow-receiving model of the environment in Fig. \ref{2ndFloor} for shadow rendering}
 \label{Transp}
 \vskip-10pt
\end{figure}

In order to make use of the 3D map created by the HoloLens, we use vertex-lighting technique to create a custom shadow-receiving shader. Although pixel lighting provides more details by calculating the illumination for each pixel, it is computationally expensive.
In contrast, by using vertex lighting, we calculate illumination at each vertex of a model and then interpolate the resulting values over the faces of the models, resulting in a more efficient solution. 
We apply this shader to the exported HoloLens-generated map (see an example in Fig. \ref{2ndFloor}) and enable its shadow receiving properties. This creates our transparent shadow-receiving model of the environment (see an example in Fig. \ref{Transp} for the environment model in Fig. \ref{2ndFloor}).
Finally, this model is superimposed onto the real-world  to render the shadow. 

\section{EXPERIMENTAL DESIGN}

\begin{figure*}
\centering
  \includegraphics[width=0.9\textwidth]{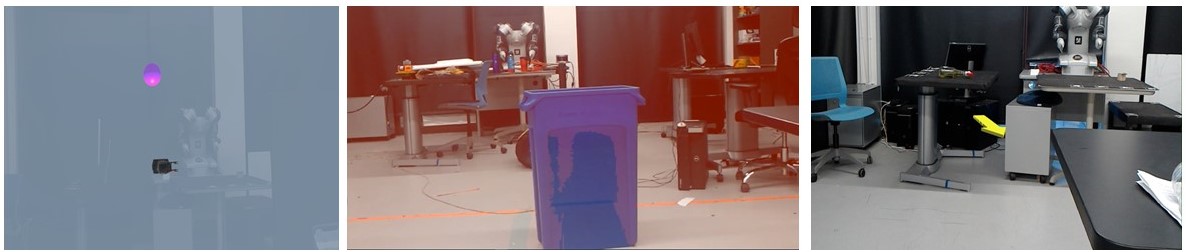}
  \caption{User view of TPS and the two baseline methods using AR: Map (left), TPS (middle), Arrow (right).}
  \label{3methods}
  \vskip-10pt
\end{figure*}

We are interested in verifying the benefits of implicit communication over explicit communication for maintaining TSA.
The focus here is on the distinction between explicit and implicit communication from the receiver's (human's) perspective. 
hence requiring less attention to receive. 
In particular, TPS is considered as implicit communication since natural shadows are not normally perceived as a way of communication by the receiver (i.e., viewer of the shadow).
In contrast, humans must pay substantial attention to receive explicit communication.
When the human has other tasks to manage simultaneously, it would result in the loss of TSA and hence reduced responsiveness to the communicator. 

To show such an effect, we compare TPS against two baselines that use explicit communication with AR.  
AR is also adopted for the baselines to avoid the impact due to differences in communication modalities.
For maintaining TSA, we assume that each method is required to {\it continuously} project team status information to the human.
TPS and the baselines using explicit communication are described below:  
\begin{enumerate}
    \item {\textbf{\textit{Map}}} shows the environment using a map. It displays a real-time view of the robot and its movements on the map (Fig. \ref{3methods} (left)). The pink sphere indicates the position of the human and the robot is shown in black. 
    \item {\textbf{\textit{Arrow}}} uses an arrow that points to the position of the robot in real-time (Fig. \ref{3methods} (right)). The arrow only pans in a plane (hence encodes no depth information). 
    \item {\textbf{\textit{TPS}}} uses a virtual shadow of the robot to communicate real-time status information (Fig. \ref{3methods} (middle)).
\end{enumerate}

Map is chosen due to its frequent use in real life,
which makes it less cognitively demanding.
It should alleviate the issue with explicit communication. 
Arrow, on the other hand, is chosen as a ``novel'' method that the human may be less familiar with. 
We chose a task scenario that is similar to the one shown in Fig. \ref{task} and our motivating example. 
In this task scenario, a human is supposed to work with a
robotic partner that occasionally delivers objects to or for the human. 
The human, on the other hand, has his own tasks (e.g., reading some documents) besides collaborating with the robot, i.e., picking up the object delivered or sending an object to be delivered. 
We used Kinova Movo in this study.

\begin{figure}[h!]
\centerline{\includegraphics[width=0.8\linewidth]{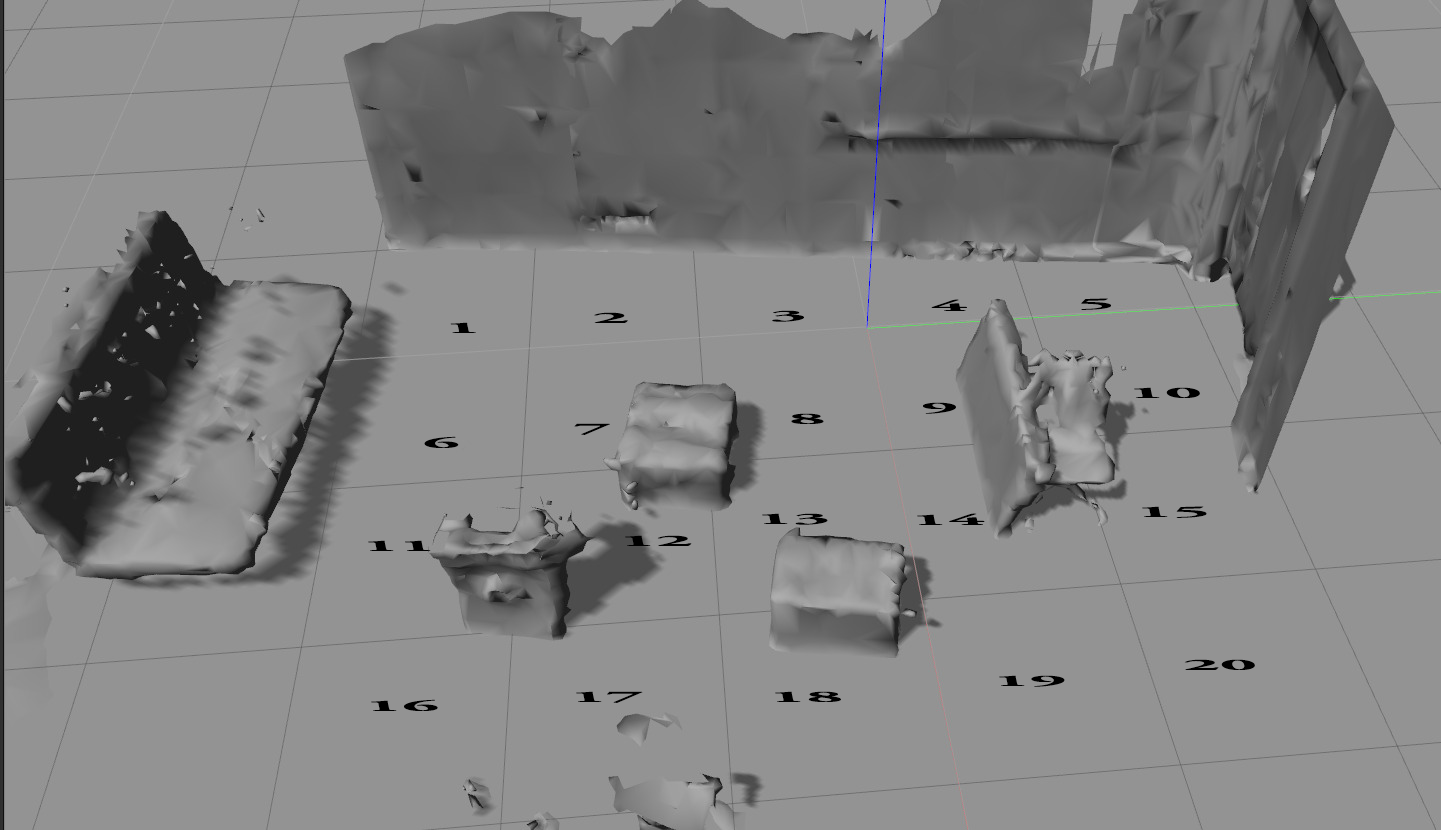}}
 \caption{Discretized environment for our study}
 \label{discrete}
 \vskip-10pt
\end{figure}

To prepare for the study, each participant was given a printed copy of the room in a discretized form for location identification (see Fig. \ref{discrete}).
We deployed all three methods onto Hololens and placed the robot outside the participant's FOV. 
To evaluate the methods for maintaining TSA, 
the participants were told they were work partners of a robot and must complete all tasks together in the least amount of time. 
The tasks included providing the robot with objects to deliver and receiving objects delivered by the robot. 
 They were asked to attend to the robot when the robot arrived to deliver or pick up an object.
At the same time, participants were asked to work on some other tasks (i.e., reading or solving simple puzzles).  
Participants were advised to not turn to observe the robot during the study and could only access the team status via Hololens. 
As a secondary aim to evaluate how accurately each method maintains TSA, 
we broke a session for each method into two parts. 
The first part involved delivery tasks as discussed
and the second part involved asking the participants to 
identify the robot's location and predict its destination (i.e., {\it estimation} tasks). 


Using the different methods, we are interested in studying how responsive the participants were to the robot (directly related to the maintenance of TSA),
how accurately their TSA was maintained, 
and how they viewed the sessions and robot after completion. 
$24$ CS students in their senior year participated in a within-subjects study. They were made up of $14$ female and $10$ male students. 
{\color{black}Each participant participated in three sessions, one for each method. We recorded a video for each part of a session (with $2$ parts in each session). The first video for the delivery tasks is used to measure the waiting time between the robot's arrival and the participant's response. The second video is recorded for the estimation tasks. 
This resulted in $6$ videos per participant and a total of $144$ videos recorded. 
However, due to objects and robot blocking the camera, or participants turning to observe the robot during the study, we discarded the data from $7$ participants, which left us with the remaining $17$ participants for result analysis. 
}
Participant were given a survey at the end of their sessions. It included an AttrakDiff survey and their preference towards the robot as a work partner in the different methods and the methods themselves w.r.t. their naturalness and user friendliness. 
In the AttrakDiff survey, the participants were asked to rate the methods based on different qualitative metrics on a scale of $1$ to $7$.
Overall, the study is designed to verify the following hypotheses:
\begin{itemize}
    \item {\it H1. TPS improves responsiveness to the robot compared to baselines using explicit communication.
    \item H2. TPS maintains accurate TSA that is comparable to the baselines using explicit communication.
    \item H3. Robot with TPS is viewed more favorably as a work partner than baselines using explicit communication. }
\end{itemize}

\section{RESULTS AND ANALYSES}

{\color{black} An alpha level of $0.005$ is used for all statistical tests.}


\subsubsection{Responsiveness to Robot}
\begin{figure}[t]
    \centering
  \includegraphics[width=0.8\linewidth]{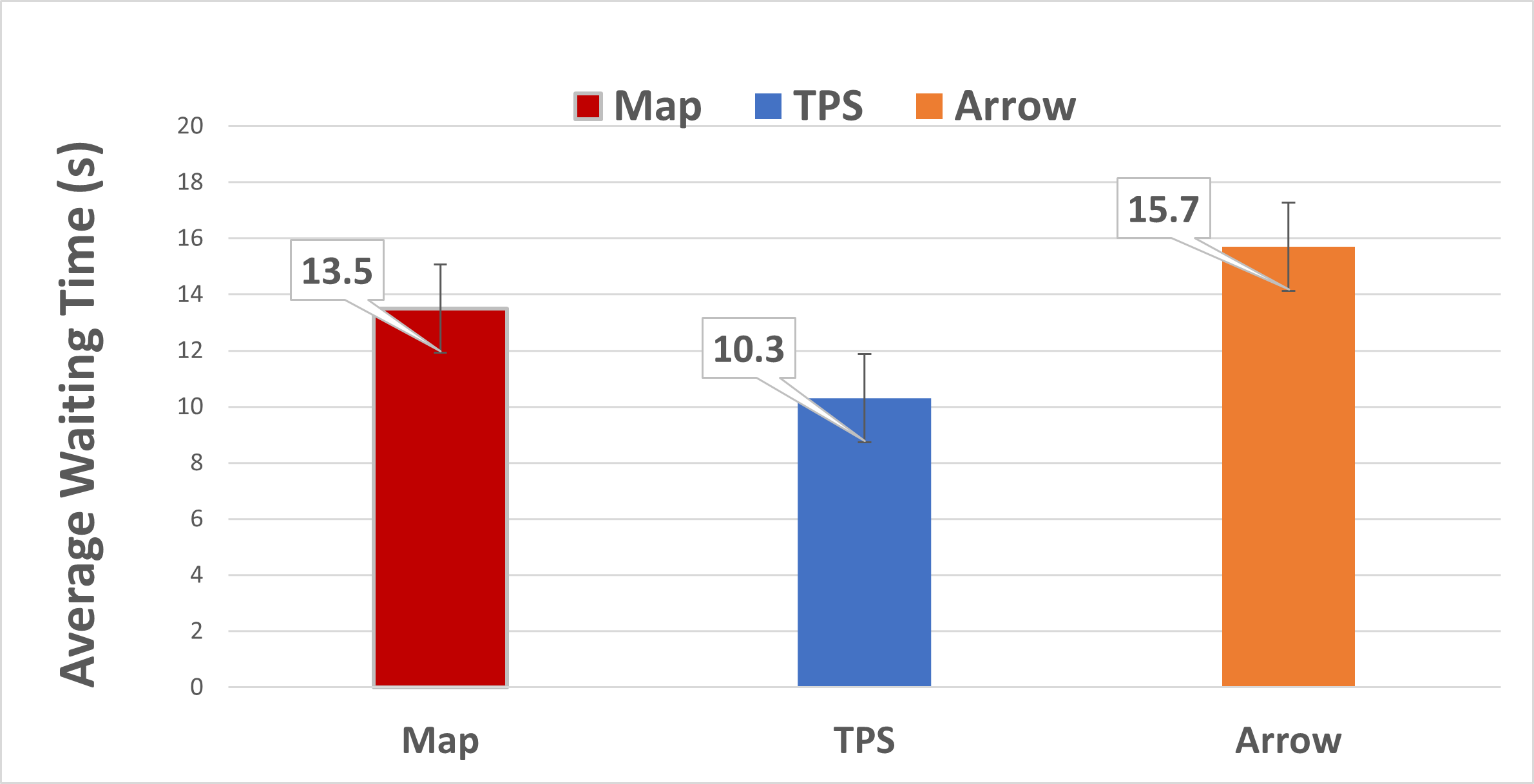}
  \caption{Waiting times for robot waiting for delivery}
  \label{WaitingTimes4RobotDElivery}
  \vskip-20pt
\end{figure}
 
\begin{figure}[t]
    \centering
  \includegraphics[width=0.8\linewidth]{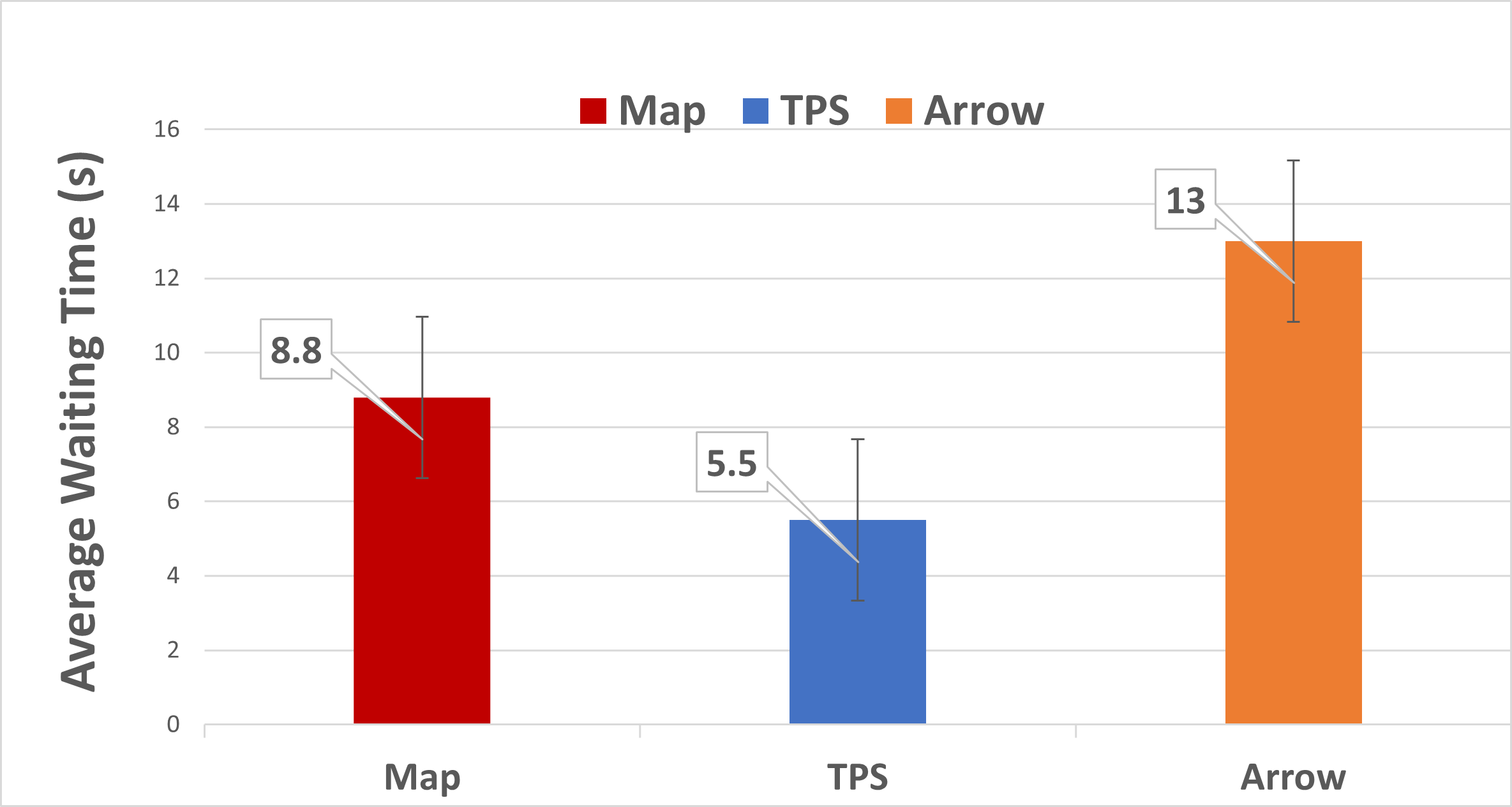}
  \caption{Waiting times for robot waiting for pickup}
  \label{waitingTimes4RobotPickUp}
  \vskip-20pt
\end{figure}

Fig. \ref{WaitingTimes4RobotDElivery} presents the results for the waiting time of the robot between when it arrived at the delivery location and when the participant responded. 
We observe that participants were much faster to react to the robot when TPS was used. 
{\color{black} 
Paired Student's t-tests gave \textit{t(16)=2.12, p=.003} between TPS (\textit{M=10.3, SD=9.18}) and Map (\textit{M=13.5, SD=7.9}) and \textit{t(16)=2.12, p=.002} between TPS (\textit{M=10.3, SD=9.18}) and Arrow (\textit{M=15.7, SD=8.8}).
Fig. \ref{waitingTimes4RobotPickUp} presents the results for the waiting time of the robot between when it arrived at the pickup location on the participant's request for a delivery and when the participant responded.
Similar results were observed.
T-tests resulted in $t(16)=2.13, p=.003$ between TPS (\textit{M=5.50, SD=2.11)} and Map (\textit{M=8.80, SD=3.62)}, 
and \textit{t(16)=1.75, p $<$ .001} between TPS (\textit{M=5.50, SD=2.11)} and Arrow (\textit{M=13.04, SD=4.66)}.
}
These results verify that the participants are more responsive to the robot in TPS than the baselines ($H1$).

\subsubsection{Accurate TSA}
\begin{figure}[h!]
    \centering
  \includegraphics[width=0.8\linewidth]{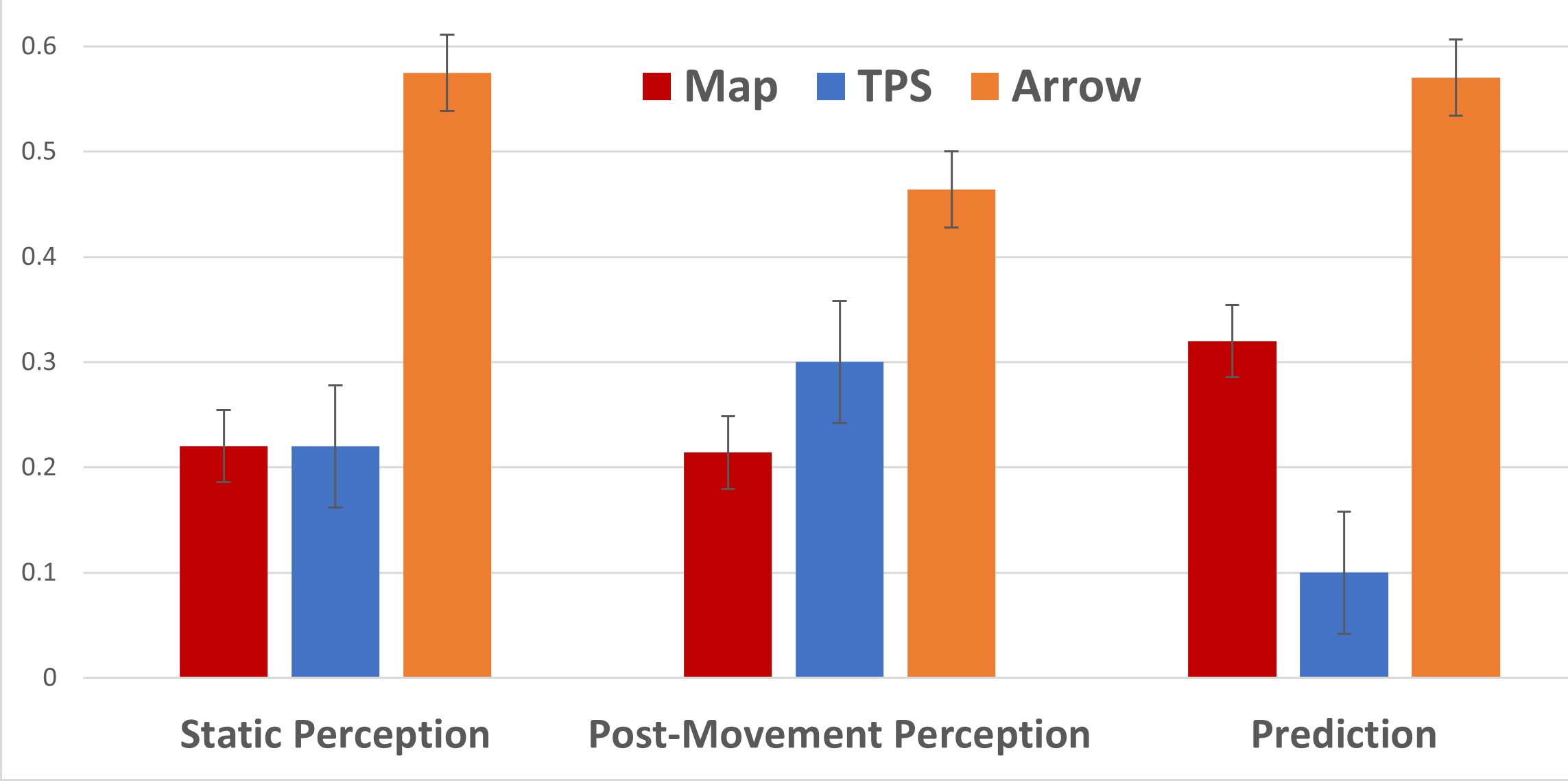}
  \caption{Comparing accuracy in estimation tasks}
  \label{stat-pred}
\end{figure}

Fig. \ref{stat-pred} presents the results w.r.t. how accurately each method maintained TSA via the estimation tasks. 
They included asking the participant to estimate where the robot's current position was ({\it static perception}), where the robot was after some movement ({\it post-movement perception}), and where they expected the destination of the robot to be ({\it prediction}). 
{\color{black} We consider the accuracy as the Manhattan distance between the participant's estimation and the ground truth. 
Hence, the lower the value, the more accurate the estimation. 
}
Generally, TPS did better than Arrow and was on a par with Map.
Map did well, which was likely due to the fact that participants were generally familiar with maps in one form or another in real life. 
Arrow performed the worst as expected since it
offered no depth information,
which made it difficult for the participants to accurately estimate the robot's position and change in position. 
It can be seen from Fig. \ref{stat-pred} that TPS performed comparably with Map in {\it perceptions}, with Map having a slight edge in post-movement perception.
{\color{black} T-tests reveal no significant differences between TPS and Map ({\it H2}).}
It is however interesting to note that TPS proved to provide more context information for TSA in {\it prediction} and did much better there than the other methods. For our prediction values (as shown on Fig. \ref{stat-pred},
We attributed such performance to better TSA since {\it prediction} required the participants to maintain the context of movements (i.e.,  which direction the robot was heading), instead of solely its position. 
{\color{black} Student's paired t-tests on only {\it prediction}  resulted in {\textit t(16)=2.19}, $p$ $<$ $.001$ between TPS (\textit{M =0.24, SD =0.42}) and Arrow (\textit{M =0.57, SD =1.08})), and \textit{t(16)=2.12, p $<$ .001} between TPS and Map (\textit{M =0.32, SD =0.55}).
}
\vskip-10pt
\vskip-15pt

\subsubsection{Attractiveness}
\begin{figure}[h!]
\centering
  \includegraphics[width=0.8\linewidth=]{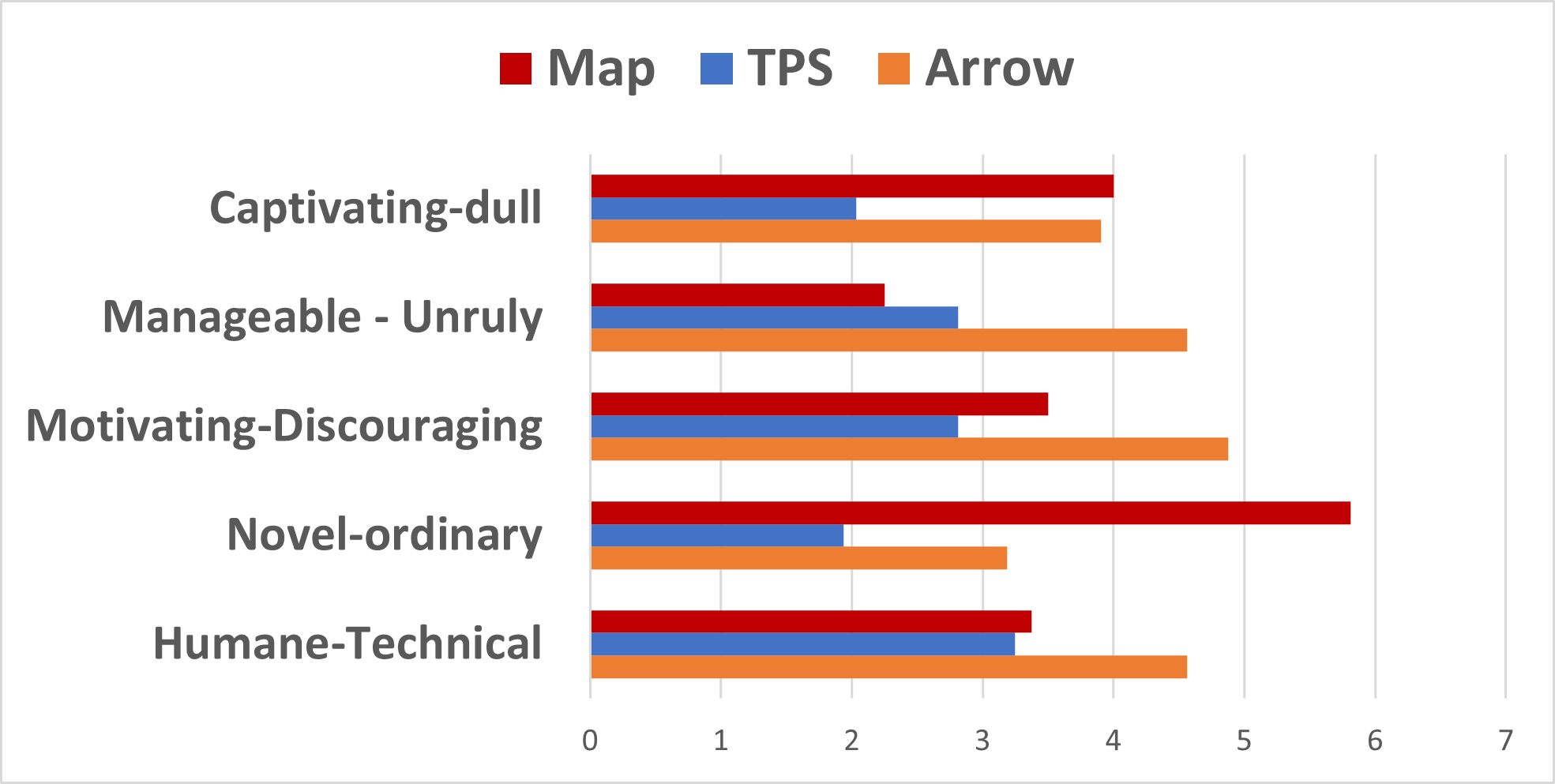}
  \caption{AttrakDiff survey results}
  \label{attrak}
  \vskip-20pt
\end{figure}
AttrakDiff evaluated the attractiveness of the robot in different methods. 
The result is presented in Fig. \ref{attrak}. 
Results indicate that the robot in TPS was viewed as more attractive than the robot in the baselines ($H3$).
It is observed that TPS obtained the best ratings among almost all features, 
although participants indicated that TPS was slightly less manageable than Map. {\color{black}We averaged the values for a Student's paired t-test.
The results are  \textit{t(16)=2.12, p $<$ .001} between TPS (\textit{M=2.59, SD=0.84}) and Map (\textit{M=3.76, SD=0.60}), and \textit{t(16)=2.13, p $<$ .001} between TPS (\textit{M=2.59, SD=0.84}) and Arrow (\textit{M=4.31, SD=0.79}).}


\begin{figure}[h!]
    \centering
  \includegraphics[width=0.8\linewidth=]{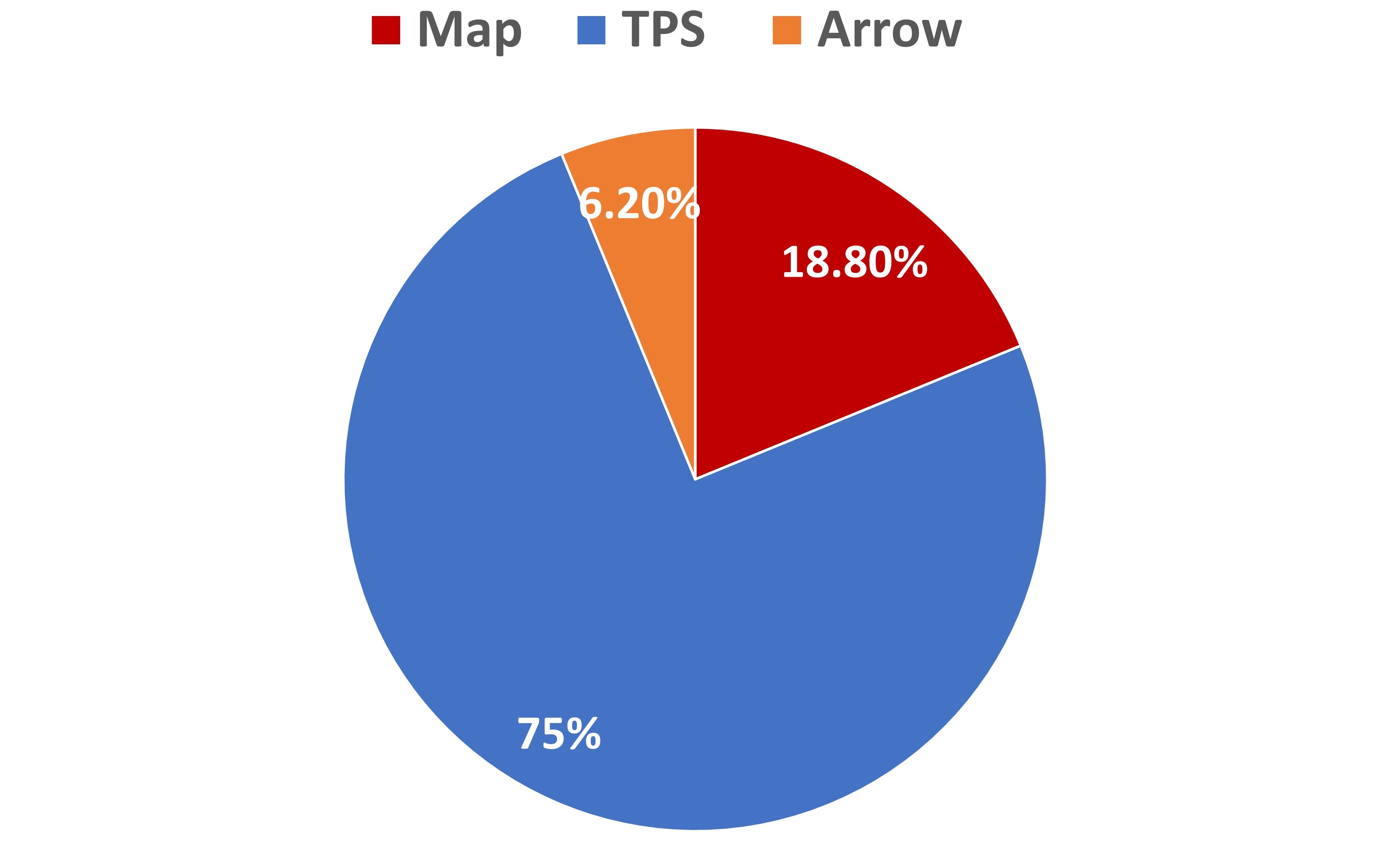}
  \caption{Participants' feeling of the robot as a work partner.}
  \label{intimate}
  \vskip-10pt
\end{figure}

\subsubsection{Partnership and Naturalness}
We explicitly asked the participants to indicate which of the three methods gave them a feeling of partnership. $75\%$ of the participants indicated that TPS gave them the feeling that the robot was a work partner. $18.8\%$ felt towards Map and only $6.2\%$ towards Arrow. Fig. \ref{intimate} shows the results. 
This result verifies that the robot with TPS would be viewed more favorably as a work partner ($H3$).
Finally, we asked the participants to indicate which methods they felt the most natural. $50\%$ answered Map while the remaining $50\%$ felt towards TPS.


\section{CONCLUSIONS}
In this paper, we introduced implicit and naturalistic team status projection via virtual shadows (TPS),
with the motivation to improve TSA and responsiveness to robots in proximal HRI. 
We addressed the challenges in realizing TPS in four steps: 
shadow mapping, shadow projection, shadow smoothing, and shadow rendering, 
resulting in a framework that can be leveraged to use virtual shadows to communicate critical information. 
TPS differs from explicit communication methods since the receiver would not normally consider such shadows as a way of communication, 
resulting in less cognitive and attention demands and better TSA. 
We evaluated TPS with two baselines that use explicit communication to demonstrate the effectiveness of our approach. 



\bibliographystyle{unsrt}
\bibliography{references}

\begin{thebibliography}{10}

\bibitem{OLIFF}
Harley Oliff, Ying Liu, Maneesh Kumar, Michael Williams, and Michael Ryan.
\newblock Reinforcement learning for facilitating human-robot-interaction in
  manufacturing.
\newblock {\em Journal of Manufacturing Systems"}, 56:326 -- 340, 2020.

\bibitem{nancy}
Nancy~J. Cooke, Jamie~C. Gorman, Christopher~W. Myers, and Jasmine~L. Duran.
\newblock Interactive team cognition.
\newblock {\em Cognitive Science}, 37(2):255--285, 2013.

\bibitem{mavridis2015review}
Nikolaos Mavridis.
\newblock A review of verbal and non-verbal human--robot interactive
  communication.
\newblock {\em Robotics and Autonomous Systems}, 63:22--35, 2015.

\bibitem{gleeson2013gestures}
Brian Gleeson, Karon MacLean, Amir Haddadi, Elizabeth Croft, and Javier
  Alcazar.
\newblock Gestures for industry intuitive human-robot communication from human
  observation.
\newblock In {\em 2013 8th ACM/IEEE International Conference on Human-Robot
  Interaction (HRI)}, pages 349--356. IEEE, 2013.

\bibitem{fong2003collaboration}
Terrence Fong, Charles Thorpe, and Charles Baur.
\newblock Collaboration, dialogue, human-robot interaction.
\newblock In {\em Robotics research}, pages 255--266. Springer, 2003.

\bibitem{cohen1995communicative}
Philip~R Cohen and Hector~J Levesque.
\newblock Communicative actions for artificial agents.
\newblock In {\em ICMAS}, volume~95, pages 65--72. Citeseer, 1995.

\bibitem{dragan2013legibility}
Anca~D Dragan, Kenton~CT Lee, and Siddhartha~S Srinivasa.
\newblock Legibility and predictability of robot motion.
\newblock In {\em 2013 8th ACM/IEEE International Conference on Human-Robot
  Interaction (HRI)}, pages 301--308. IEEE, 2013.

\bibitem{zhang2017plan}
Yu~Zhang, Sarath Sreedharan, Anagha Kulkarni, Tathagata Chakraborti,
  Hankz~Hankui Zhuo, and Subbarao Kambhampati.
\newblock Plan explicability and predictability for robot task planning.
\newblock In {\em 2017 IEEE international conference on robotics and automation
  (ICRA)}, pages 1313--1320. IEEE, 2017.

\bibitem{reis2011familiarity}
Harry~T Reis, Michael~R Maniaci, Peter~A Caprariello, Paul~W Eastwick, and
  Eli~J Finkel.
\newblock Familiarity does indeed promote attraction in live interaction.
\newblock {\em Journal of personality and social psychology}, 101(3):557, 2011.

\bibitem{boateng2021virtual}
Andrew Boateng and Yu~Zhang.
\newblock Virtual shadow rendering for maintaining situation awareness in
  proximal human-robot teaming.
\newblock In {\em Companion of the 2021 ACM/IEEE International Conference on
  Human-Robot Interaction}, pages 494--498, 2021.

\bibitem{aph_144559232}
Igor Baranovski, Stevan Stankovski, Gordana Ostojic, Sabolč Horvat, and Srdjan
  Tegeltija.
\newblock Augmented reality support for self-service automated systems.
\newblock {\em Journal of graphic engineering and design}, 11:63--68, 06 2020.

\bibitem{Mekni2014AugmentedR}
Mehdi Mekni and Andre Lemieux.
\newblock Augmented reality: Applications, challenges and future trends.
\newblock {\em Applied computational science}, 20:205--214, 2014.

\bibitem{liv}
M.~A. Livingston.
\newblock An augmented reality system for military operations in urban terrain.
\newblock {\em I/ITSEC2002, Dec.}, 2002.

\bibitem{marketing}
Julien Pilet, Vincent Lepetit, and Pascal Fua.
\newblock Fast non-rigid surface detection, registration and realistic
  augmentation.
\newblock {\em International Journal of Computer Vision}, 76, 02 2008.

\bibitem{MustafaS}
Mustafa Sirakaya and Didem~Alsancak Sirakaya.
\newblock Trends in educational augmented reality studies: A systematic review.
\newblock {\em Malaysia online journal of educational technology}, 6(2):60--74,
  2018.

\bibitem{engineering}
S.J Henderson and S~Feiner.
\newblock Evaluating the benefits of augmented reality for task localization in
  maintenance of an armored personnel carrier turret.
\newblock {\em ISMAR}, pages 135--144, 2009.

\bibitem{medical}
Carolien Kamphuis, Esther Barsom, Marlies Schijven, and Noor Christoph.
\newblock Augmented reality in medical education?
\newblock {\em Perspectives on medical education}, 3(4):300—311, September
  2014.

\bibitem{entertainment}
Yolanda Vazquez-Alvarez, Ian Oakley, and Stephen~A Brewster.
\newblock Auditory display design for exploration in mobile audio-augmented
  reality.
\newblock {\em Personal and Ubiquitous Computing}, 16:987--999, 2012.

\bibitem{robot}
George Michalos, Panagiotis Karagiannis, Sotiris Makris, Önder Tokçalar, and
  George Chryssolouris.
\newblock Augmented reality (ar) applications for supporting human-robot
  interactive cooperation.
\newblock {\em Procedia CIRP}, 41:370 -- 375, 2016.
\newblock Research and Innovation in Manufacturing: Key Enabling Technologies
  for the Factories of the Future - Proceedings of the 48th CIRP Conference on
  Manufacturing Systems.

\bibitem{WANG2003185}
Yang Wang and Dimitris Samaras.
\newblock Estimation of multiple directional light sources for synthesis of
  augmented reality images.
\newblock {\em Graphical Models}, 65(4):185 -- 205, 2003.

\bibitem{control}
Ayoung Hong, Burak Zeydan, Samuel Charreyron, Olgac Ergeneman, Salvador Pane,
  M.~Toy, Andrew Petruska, and Brad Nelson.
\newblock Real-time holographic tracking and control of microrobots.
\newblock {\em IEEE Robotics and Automation Letters}, 2:1--1, 01 2017.

\bibitem{heni}
R.~S. {Andersen}, O.~{Madsen}, T.~B. {Moeslund}, and H.~B. {Amor}.
\newblock Projecting robot intentions into human environments.
\newblock In {\em 2016 25th IEEE International Symposium on Robot and Human
  Interactive Communication (RO-MAN)}, pages 294--301, 2016.

\bibitem{carston2008thoughts}
Robyn Carston.
\newblock {\em Thoughts and utterances: The pragmatics of explicit
  communication}.
\newblock John Wiley \& Sons, 2008.

\bibitem{carston2009explicit}
Robyn Carston.
\newblock The explicit/implicit distinction in pragmatics and the limits of
  explicit communication.
\newblock {\em International review of pragmatics}, 1(1):35--62, 2009.

\bibitem{breazeal2005effects}
Cynthia Breazeal, Cory~D Kidd, Andrea~Lockerd Thomaz, Guy Hoffman, and Matt
  Berlin.
\newblock Effects of nonverbal communication on efficiency and robustness in
  human-robot teamwork.
\newblock In {\em 2005 IEEE/RSJ international conference on intelligent robots
  and systems}, pages 708--713. IEEE, 2005.

\bibitem{che2020efficient}
Yuhang Che, Allison~M Okamura, and Dorsa Sadigh.
\newblock Efficient and trustworthy social navigation via explicit and implicit
  robot--human communication.
\newblock {\em IEEE Transactions on Robotics}, 36(3):692--707, 2020.

\bibitem{unity}
Microsoft.
\newblock Unity - manual: Shadows.
\newblock 2015.

\bibitem{PIDforRos}
Manuel Beschi, Riccardo Adamini, Alberto Marini, and Antonio Visioli.
\newblock Using of the robotic operating system for pid control education.
\newblock {\em IFAC-PapersOnLine}, 48(29):87--92, 2015.
\newblock IFAC Workshop on Internet Based Control Education IBCE15.

\end{thebibliography}

\end{document}